\documentclass[sigconf]{acmart}
\usepackage{multirow}
\usepackage{subfigure}
\usepackage{makecell}

\AtBeginDocument{%
  \providecommand\BibTeX{{%
    \normalfont B\kern-0.5em{\scshape i\kern-0.25em b}\kern-0.8em\TeX}}}

\setcopyright{acmlicensed}
\copyrightyear{2018}
\acmYear{2018}
\acmDOI{XXXXXXX.XXXXXXX}

\acmConference[MM'24]{Make sure to enter the correct
  conference title from your rights confirmation email}{October 28 - November 1,
  2024}{Melbourne, Australia.}
\acmISBN{978-1-4503-XXXX-X/18/06}

\begin{document}

\title{Fake News Detection and Manipulation Reasoning via Large Vision-Language Models}


\author{Ruihan Jin}
\affiliation{%
  \institution{Department of Automation, Tsinghua University}
  \country{China}}
\email{jrh20@mails.tsinghua.edu.cn}

\author{Ruibo Fu$^\dag$}
\affiliation{%
  \institution{Institute of Automation, Chinese Academy of Sciences}
  \thanks{$\dag$ Corresponding author}
  \country{China}}
\email{ruibo.fu@nlpr.ia.ac.cn}

\author{Zhengqi Wen}
\affiliation{%
  \institution{Institute of Automation, Chinese Academy of Sciences}
  \country{China}}
\email{zqwen@nlpr.ia.ac.cn}

\author{Shuai Zhang}
\affiliation{%
  \institution{Department of Automation, Tsinghua University}
  \country{China}}
\email{zhang_shuai@mail.tsinghua.edu.cn}

\author{Yukun Liu}
\affiliation{%
  \institution{School of Artificial Intelligence, University of Chinese Academy of Sciences}
  \country{China}}
\email{yukunliu927@gmail.com}

\author{Jianhua Tao}
\affiliation{%
  \institution{Department of Automation, Tsinghua University}
  \country{China}}
\email{jhtao@tsinghua.edu.cn}
\renewcommand{\shortauthors}{Ruihan Jin et al.}

\begin{abstract}
  Fake news becomes a growing threat to information security and public opinion with the rapid sprawl of media manipulation. Therefore, fake news detection attracts widespread attention from academic community. Traditional fake news detection models demonstrate remarkable performance on authenticity binary classification but their ability to reason detailed faked traces based on the news content remains under-explored. Furthermore, due to the lack of external knowledge, the performance of existing methods on fact-related news is questionable, leaving their practical implementation unclear. In this paper, we propose a new multi-media research topic, namely manipulation reasoning. Manipulation reasoning aims to reason manipulations based on news content. To support the research, we introduce a benchmark for fake news detection and manipulation reasoning, referred to as Human-centric and Fact-related Fake News (HFFN). The benchmark highlights the centrality of human and the high factual relevance, with detailed manual annotations. HFFN encompasses four realistic domains with fake news samples generated through three manipulation approaches. Moreover, a Multi-modal news Detection and Reasoning langUage Model (M-DRUM) is presented not only to judge on the authenticity of multi-modal news, but also raise analytical reasoning about potential manipulations. On the feature extraction level, a cross-attention mechanism is employed to extract fine-grained fusion features from multi-modal inputs. On the reasoning level, a large vision-language model (LVLM) serves as the backbone to facilitate fact-related reasoning. A two-stage training framework is deployed to better activate the capacity of identification and reasoning. Comprehensive experiments demonstrate that our model outperforms state-of-the-art (SOTA) fake news detection models and powerful LVLMs like GPT-4 and LLaVA.
\end{abstract}





\maketitle

\begin{figure}[t]
    \centering
    \includegraphics[scale=0.16]{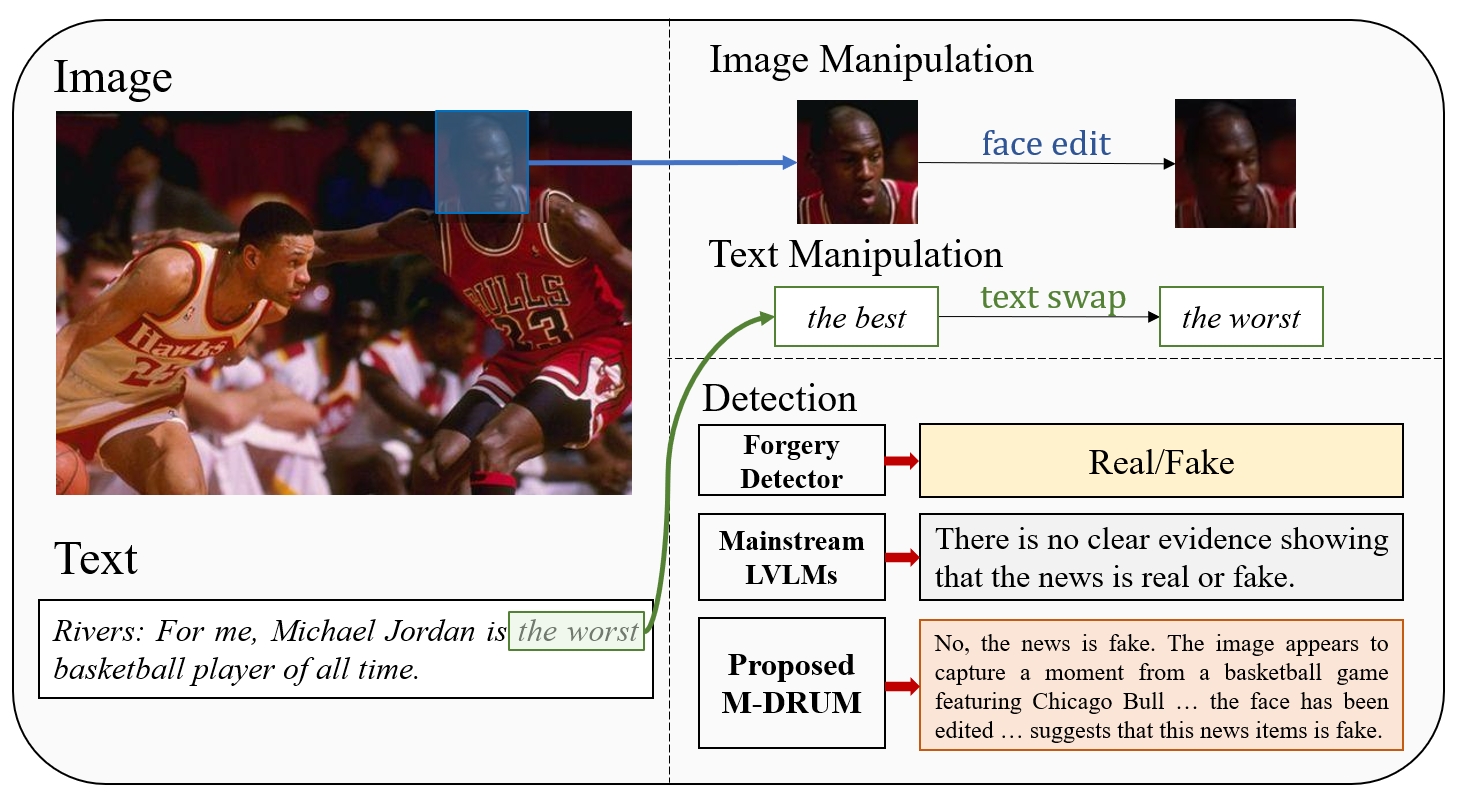}
    \caption{An illustration of multi-modal fake new detection and manipulation reasoning. We construct \textbf{H}uman-centric and \textbf{F}act-related \textbf{F}ake \textbf{N}ews(\textbf{HFFN}) benchmark through three approaches of media manipulation. We proposed \textbf{M}ulti-modal news \textbf{D}etection and \textbf{R}easoning lang\textbf{U}age \textbf{M}odel(\textbf{M-DRUM}) to not only perform authenticity classification but also reason about manipulations.}
    \label{fig:overview}
\end{figure}

\section{Introduction}

The development of online media greatly improves the convenience of information communication. On the contrast, recent years witness a rampancy of disinformation, which poses threat to information security and public opinion. News is at enormous risk of manipulation for being a common carrier of multi-modal information, which draws attention within the academia community and various fake news detection methods are proposed. Early works on fake news detection prioritize the identification of uni-modal manipulation. Currently, with the advent of deep generative models, media manipulation expands across multiple modalities. Visual deepfake models can edit human faces and generate high-fidelity images and videos~\cite{patashnik2021styleclip,wang2022high}. With large language models (LLMs) like BERT~\cite{devlin2018bert} and GPT~\cite{chatgpt}, lexical replacement and editing is performed easily to modify semantics and facts. Media manipulation enhances the difficulty of detection and has a more detrimental social impact when it targets human-centric and fact-related news.

To address multi-modal media manipulation, modern fake news detection approaches leverage feature-level interaction of different modalities~\cite{chen2022cross,wu2021multimodal,shao2023detecting}. Despite the favourable performance achieved, two major challenges still exist. First, most forgery detectors fail to reason about potential manipulations. Mere authenticity binary classification is trivial for analyzing manipulations or sorting out forgery mechanism, which limits the practical implementation. Second, as mentioned above, manipulations tend to attack human-centric news involving celebrities or well-known events, with a high factual relevance. It is critical to identify human-centric and fact-related fake news to eliminate negative social impacts.

In this paper,a new multi-media research task is proposed, namely manipulation reasoning. Manipulation reasoning aims to reason about potential manipulations based on news content. Existing benchmarks failed to provide analytical reasoning about the manipulation on news and lack the bias toward human-centric and fact-related news. To facilitate further research, we present a benchmark for fake news detection and manipulation reasoning, which is designed for both forgery detectors and general-purposed LVLMs. The benchmark is referred to as \textbf{H}uman-centric and \textbf{F}act-related \textbf{F}ake \textbf{N}ews(\textbf{HFFN}). Specifically, HFFN collects multi-modal news represented by image-text pairs, encompassing four domains: \textit{entertainment}, \textit{sport}, \textit{politics} and \textit{others}. News samples in HFFN emphasize the centrality of human and high factual relevance. Three manipulation approaches are developed to perform multi-modal fake news generation. Furthermore, detailed manual annotations are attached to news to facilitate manipulation reasoning. 

As illustrated in Fig.\ref{fig:overview}, neither existing fake news detection models nor mainstream LVLMs achieve satisfactory results on multi-modal news. Performing fake news detection and manipulation reasoning urges a combination of authenticity representation and general knowledge. Owning a wealth of general knowledge, large vision-language models (LVLMs) are cut out for manipulation reasoning. In this paper, we proposed \textbf{M}ulti-modal news \textbf{D}etection and \textbf{R}easoning lang\textbf{U}age \textbf{M}odel(\textbf{M-DRUM}), a novel large vision-language model for multi-modal fake news detection. M-DRUM aligns images and texts with a multi-modal encoder and leverages a manipulation-specific facial feature to enhance human-centric representation. Multi-modal features are aggregated through cross-modal fusion and are prompted to an LVLM to generate detection results and analytical reasoning. To our best knowledge, we are the first to employ LVLM as the backbone model for fake news detection. Comprehensive experimental results demonstrate that M-DRUM outperforms SOTA multi-modal fake news detection models and mainstream LVLMs. The enhancement of few-shot learning and chain-of-thought reasoning is confirmed with further experiments.

Main contributions of this paper are as follows:

\begin{itemize}
    \item We present the fake news detection and manipulation reasoning benchmark HFFN. The benchmark is constructed following the principle of "\textit{human-centric}" and "\textit{fact-related}". Fake news samples in HFFN are generated through three manipulation approaches and encompass four realistic domains.
    \item We propose M-DRUM, a novel large vision-language model for fake news detection and manipulation reasoning. M-DRUM not only detect authenticity classes based on the multi-modal news, but also perform analytical reasoning about potential manipulations.
    \item Comprehensive experiments demonstrate M-DRUM outperforms SOTA multi-modal fake news detection models and powerful LVLMs like GPT-4 and LLaVA both quantitatively and qualitatively. Further experiments verify the improvement of few-shot learning and chain-of-thought reasoning.
\end{itemize}

\section{Related Work}

\subsection{Media Manipulation.}

Disinformation becomes a growing threat to information security and public opinion with the rampancy of media manipulation. Media manipulation methods varies across different modalities. In visual modality, GAN-based methods are widely employed to manipulate human faces with text-guidance~\cite{nam2018text,liu2020describe} or latent space editing~\cite{richardson2021encoding,tov2021designing}. ~\cite{patashnik2021styleclip} utilizes the multi-modal semantics to guide the editing process. ~\cite{wang2022high} enables high-fidelity image inversion and attribute editing by a distortion consultation approach. In textual modality, common manipulation methods include conditional text generation~\cite{brown2020language,raffel2020exploring} and text style transfer~\cite{sheikha2011generation,sudhakar2019transforming}. Recent progress in natural language generation gives rise to large-scale manipulable text~\cite{crothers2023machine}. Manipulations toward human-centric and fact-related news may cause harmful impact to society. In our work, by applying off-the-shelf manipulation methods, we build a multi-modal fake news benchmark following the principle of "\textit{human-centric}" and "\textit{fact-related}" to evaluate detection and reasoning.

\subsection{Fake News Detection.}

Fake news detection draws great attention as news is at enormous risk of multi-modal manipulation. Social context based detection methods judge on the authenticity of news based on the spreading procedure such as social network~\cite{nguyen2020fang} and post-user interaction~\cite{min2022divide}. Content-based methods differentiate fake and real news by finding manipulation cues~\cite{przybyla2020capturing,qi2021improving}. Recent researches focus on identifying multi-modal news. ~\cite{wu2021multimodal} proposes an effective textual and visual feature fusion method with co-attention. ~\cite{chen2022cross,zhou2023multi} leverage the adaptable aggregation between uni-modal and cross-modal features to resolve the inherent ambiguity across different modalities. ~\cite{hu2023badactor} introduces LLM as a data augmentation approach to generate advisable rationales for subsequent detection. Different from aforementioned methods, we propose a novel architecture combining feature extraction and LVLM for fake news detection and manipulation reasoning. To our best knowledge, we are the first to employ LVLM as the backbone model for fake news detection.

\subsection{Large Vision-Language Models.}

Expanding the multi-modal capability of LLMs is a current research focus. ~\cite{li2023blip} employs Flan-T5~\cite{chung2022scaling} with a Q-Former to bridge the modality gap between visual feature and language model. ~\cite{su2023pandagpt} leverages the combination of ImageBind~\cite{girdhar2023imagebind} and Vicuna~\cite{vicuna2023} to deal with multi-modal input. By instruction tuning on multi-modal instruction-following data generated by GPT-4~\cite{achiam2023gpt}, ~\cite{liu2023visual,liu2023improved} achieve impressive cross-modal chat abilities. Despite the general knowledge derived from large-scale pre-training, these models lack domain-specific expertise. To better prompt LVLMs with manipulation detection expertise, we introduce a multi-level prompt learner to enhance manipulation reasoning. Fig.\ref{fig:overview} exhibits M-DRUM outperforms existing forgery detectors and LVLMs with profound manipulation detection expertise and broad general knowledge.

\section{HFFN: Human-centric and Fact-related Fake News Benchmark}

\subsection{Design Principles}

\noindent\textbf{Human-Centric} Human-centric news carries a higher risk of manipulation than other news topics. High-fidelity deepfake models can perform face swap and facial attribute editing easily, posing harmful threats to visual authenticity. Human centric news is highlighted in the construction of our benchmark, which means we pay higher attention to news samples with clear human faces. Images with no faces or blurred faces are filtered out. To simulate potential image manipulations, face swap and facial-attribute editing are both conducted to create fake images.

\noindent\textbf{Fact-Related} Factual errors are common in media manipulation, which result in misleading public opinion and negative social impact. Traditional fake news detection methods have difficulty distinguishing factual errors due to the lack of general knowledge or external knowledge source. There is a strong demand to measure whether a detection model is capable of tackling factual factual manipulation. Our benchmark is tailored for LVLMs and contains sufficient fact-related news samples. During data collection, we gather news featuring celebrities and well-known events as we believe these news is at a higher risk of being factually manipulated. After collection, random factual errors are added to the news to examine the capacity of the detection model.

\subsection{Construction Process: Data Collection}

The original news samples are gathered from latest real-world media sources. Among them, human-centric and fact-related samples get the most attention. Following the design principles, news without clear human faces or high factual relevance is screened out. The screening process is carefully conducted by multiple volunteers. To enhance validity, we calculate the image-text consistency of news by CLIP~\cite{radford2021learning}. News with low image-text consistency are removed to improve the validity of the benchmark. To this end, the original news samples set $\mathcal{O}=\{I_{real}, T_{real}\}$ is obtained.

\subsection{Construction Process: Media Manipulation}

\noindent\textbf{Image Manipulation} Inspired by manipulation procedure of ${\rm DGM}^4$, we achieve image manipulation with face swap and face editing. Face swap manipulation refers to replace the main character's face with another person's. InfoSwap~\cite{gao2021information} is adopted to swap faces by replacing the largest face $f_o$ in the original image $I_o$ with a random source face $f_{swap}$ from CelebA-HQ dataset~\cite{karras2017progressive}. Face editing manipulation refers to modify the facial attributes of the main character. For example, we intentionally put a smiling face or render an exaggerated beard on his/her face. We achieve high-fidelity editing effects with a GAN-based method~\cite{wang2022high} to transfer the original face $f_o$ into target style $f_{edit}$. In both ways, the manipulated face is stitched back to the original image $I_o$ and the manipulated image $I_s$ is get. The bounding box of the manipulated region $b=\{x_1, y_1, x_2, y_2\}$ is recorded as annotation.

\noindent\textbf{Text Manipulation} In text manipulation, the textual semantics are attacked by word substitution. Assisted by ChatGPT~\cite{chatgpt}, words in the input headline $T_o$ are revised to reverse the global semantics of the text. For example, an original headline of "Liu Xiang returns triumphantly and receives heated extolling" is altered as "Liu Xiang returns triumphantly and receives harsh questioning". Therefore, the global semantic of the headline is reversed.

\noindent\textbf{Factual Manipulation} Among various factual manipulation methods for text, entity replacement is one of the most common and convenient, which is adopted to create samples with factual errors. Specifically, given the input headline $T_o$, a Named Entity Recognition(NER) model~\cite{cui2021template} is launched to extract the name of the main entity. The extracted name is replaced with a randomly chosen name subsequently. We record the manipulated text derived from text manipulation and factual manipulation as $T_s$.

Three uni-modal manipulation approaches mentioned above are conducted on the original dataset $\mathcal{O}$ by randomly alter the original samples $I_o, T_o$ with manipulated ones $I_s, T_s$. A total of five manipulation types are formed, including three uni-modal types(Image, Text and Fact) and two cross-modal types(Image\&Text and Image\&Fact).

\begin{figure}[t]
    \subfigure[Manipulation types in HFFN.]{
    \label{fig:faketype}
    \includegraphics[width=0.47\columnwidth]{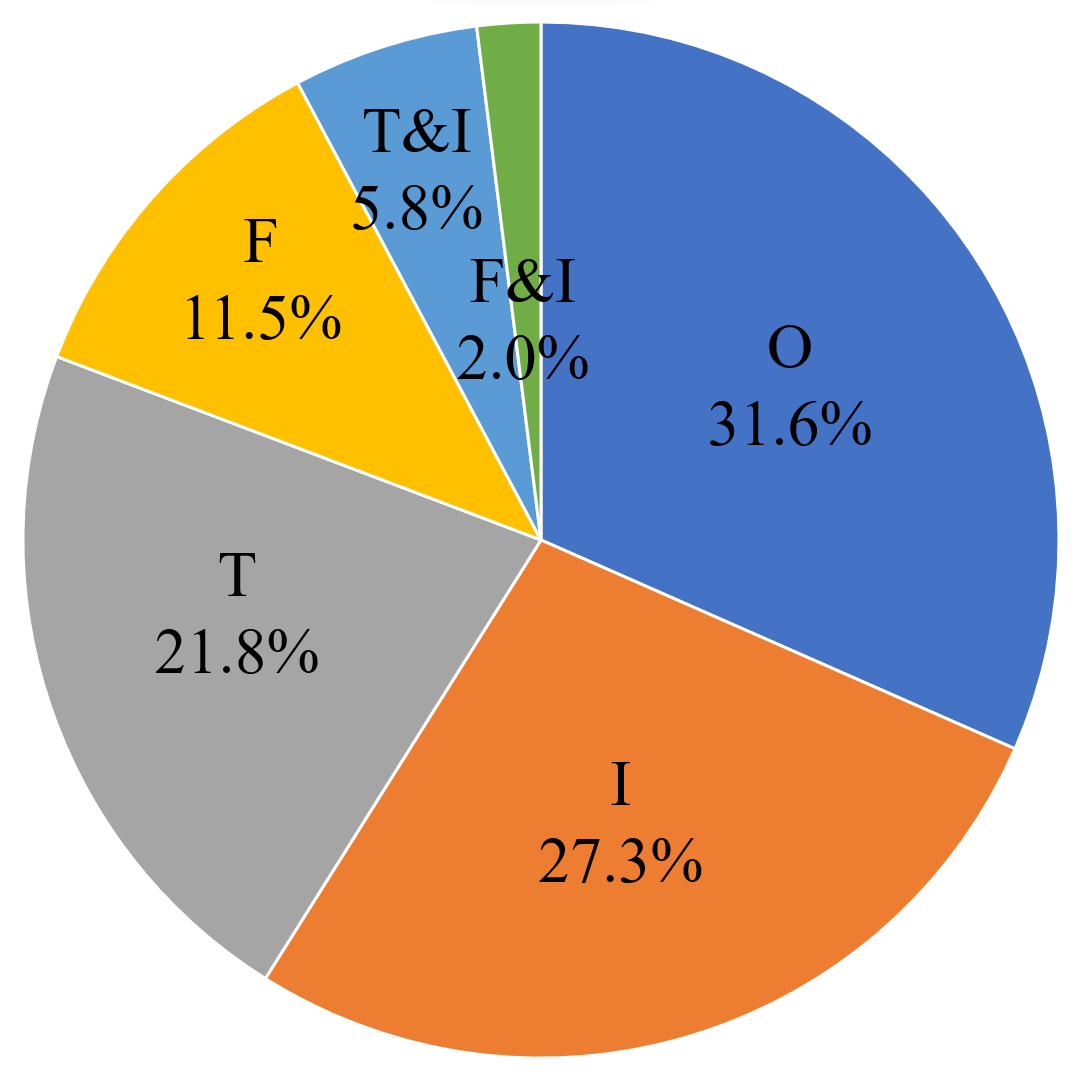}
    }
    \subfigure[Four domains in HFFN.]{
    \label{fig:domain}
    \includegraphics[width=0.47\columnwidth]{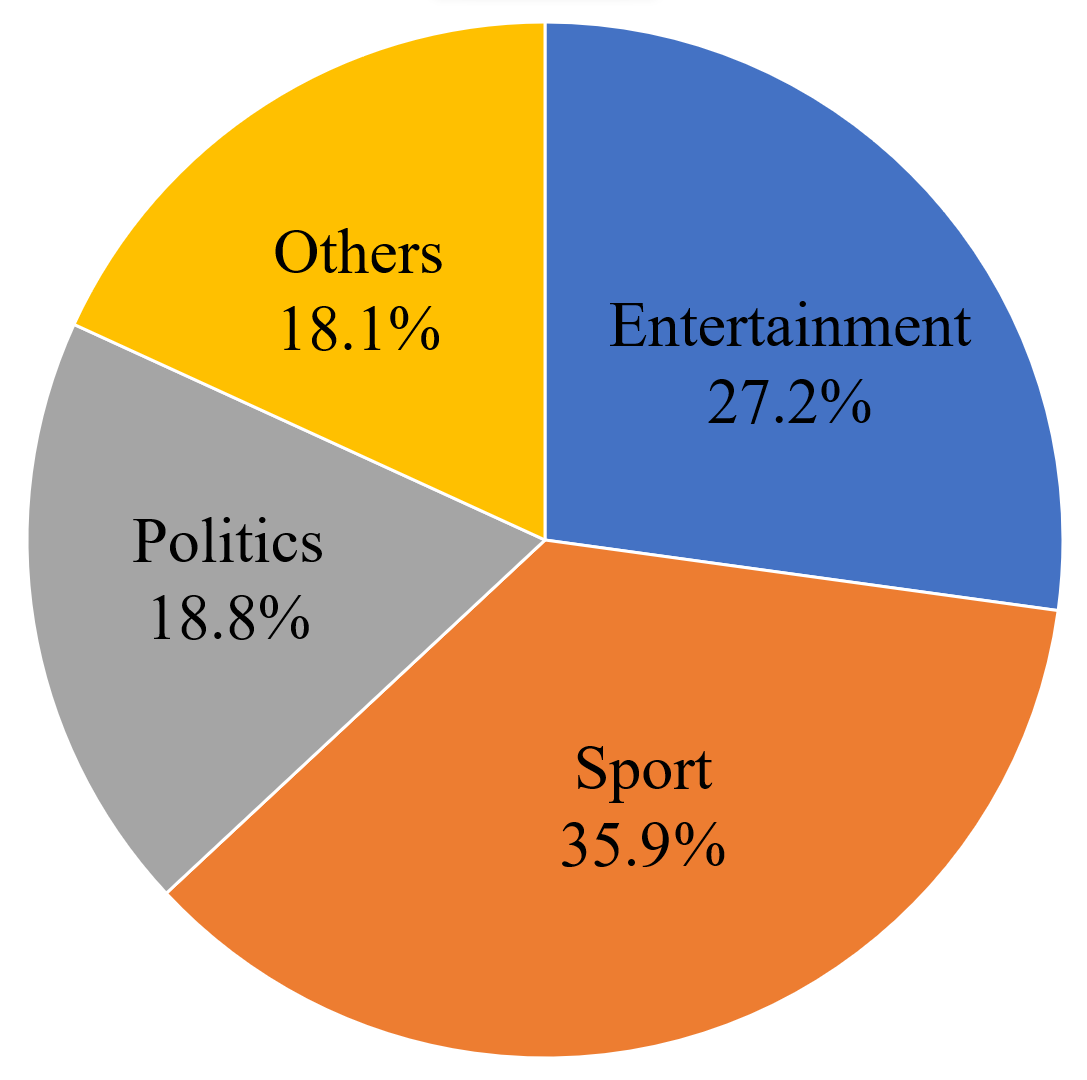}
    }	
    \subfigure[Distribution of different manipulation types]{
    \label{fig:distribution}
    \includegraphics[width=0.90\columnwidth]{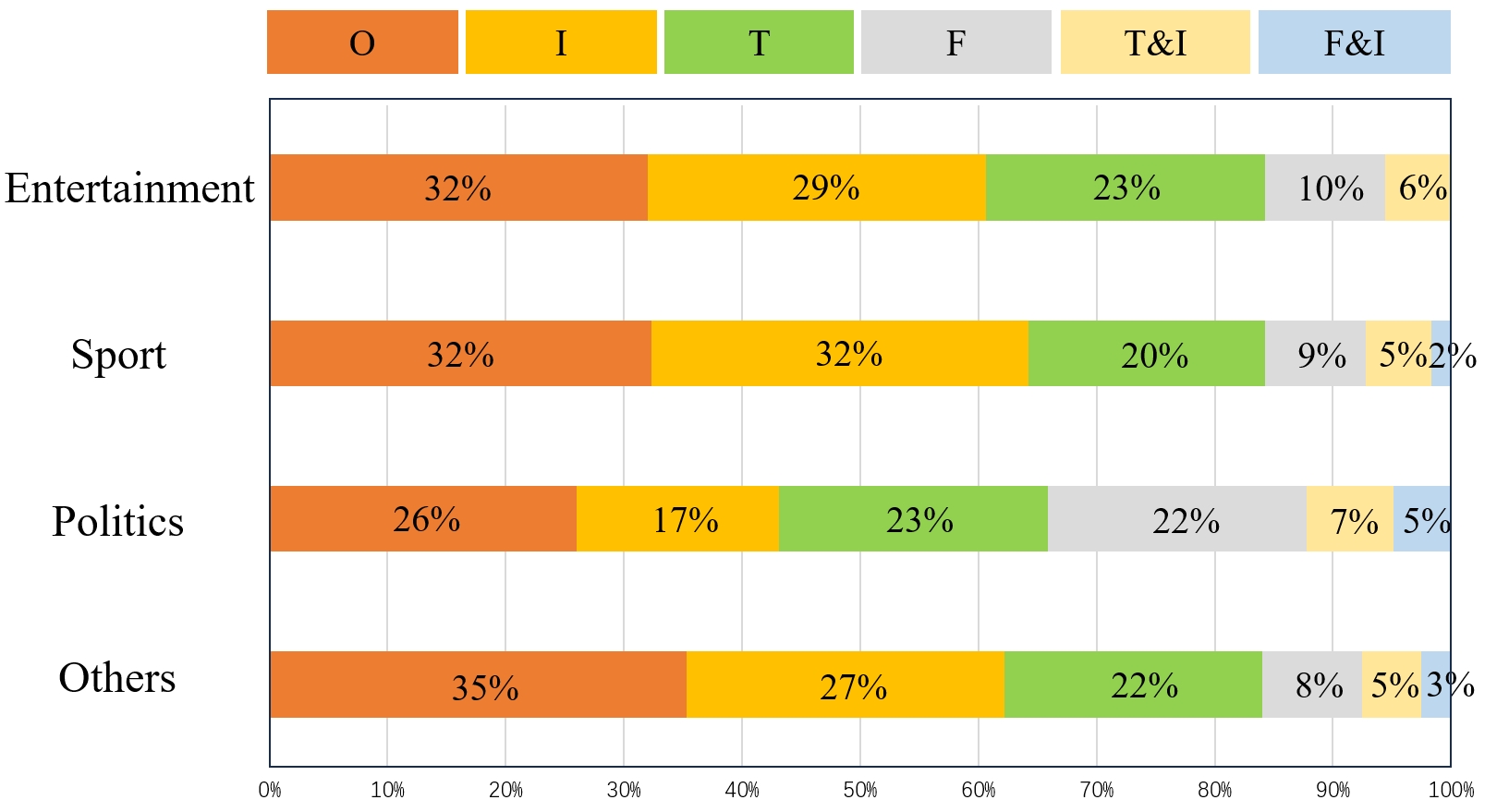}
    }
    \caption{Statistics of \textbf{HFFN} benchmark. (I: Image Manipulation, T: Text Manipulation, F: Factual Manipulation, \&: combination of two manipulation types)}
    \label{fig:sta}
\end{figure} 

\begin{figure*}[t]
    \centering
    \includegraphics[width=1\textwidth]{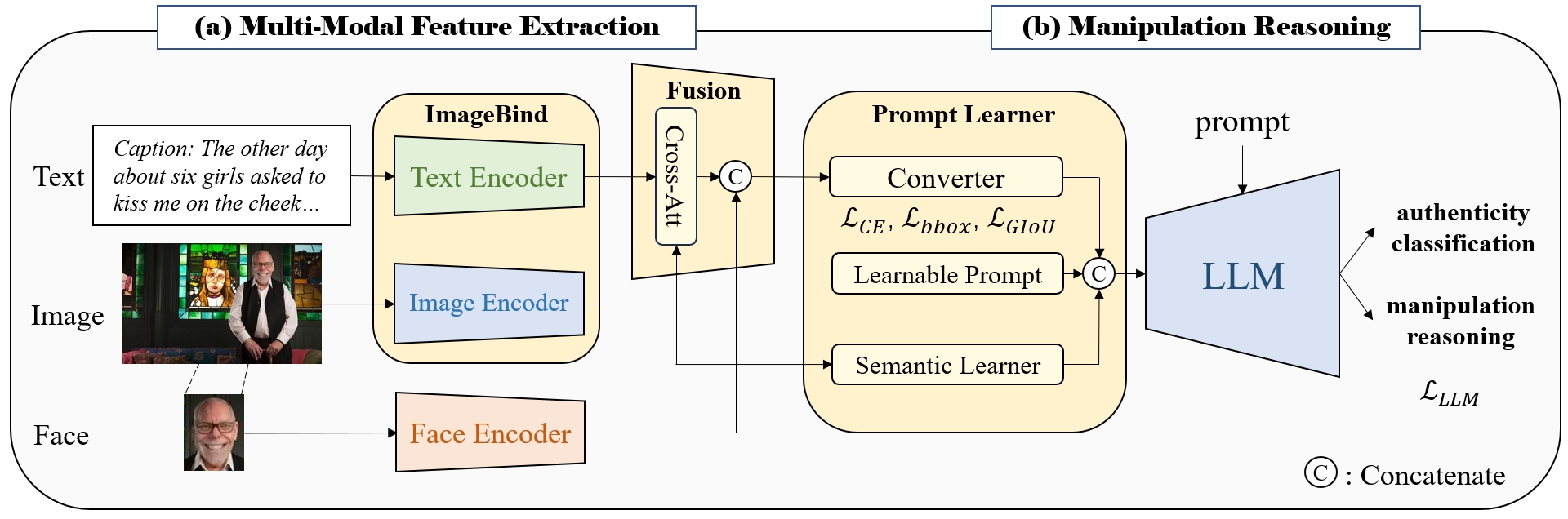}
    \caption{The architecture of M-DRUM. In M-DRUM, news images and headlines are aligned with a multi-modal encoder and a manipulation-specific facial feature is leveraged to enhance human-centric representation. Fusion features are derived with the cross-attention mechanism. To bridge the gap between the manipulation expertise and the general knowledge of LVLM, a prompt learner is adopted and a LVLM raises authenticity classification and manipulation reasoning. The model is trained under a two-stage framework to strengthen the capacity of identification and reasoning.}
    \label{fig:architecture}
\end{figure*}

\subsection{Construction Process: Human Annotation}

In HFFN benchmark, we annotate the multimodal news samples with detailed evidence for manipulation reasoning. Annotating HFFN is a challenging task, which requires annotators to endow with background knowledge in the relevant field and provide analytical reasoning based on details of news. We hire 10 professional annotators to annotate the reasoning process with following steps:

1) \textit{Authenticity annotation.} Point out the authenticity of the in terms of "Yes, the news is real" or "No, the news is fake", given the manipulation type of the news.

2) \textit{Content summary.} Raise a description of news contents for specific content analysis. The description includes the perspective of image content, headline description and image-text consistency.

3) \textit{Clue revelation.} Provide detailed clues or reasoning for unveiling manipulations such as traces of face editing or factual errors in the headline.

During annotation, we provide news contents and corresponding manipulation types for annotators and they are required to annotate the reasoning process. Each news item is shown to two annotators to perform independent annotation. Comparing the quality of their annotations, the final annotation is selected from them.

\subsection{Overview of HFFN}

The overall statistics of HFFN are visualized in Fig.\ref{fig:sta}. In line with the design principles, HFFN benchmark attaches great significance to the human-centric and fact-related news. HFFN consists a total of 655 samples, encompassing four realistic domains: \textit{entertainment}, \textit{sport}, \textit{politics} and \textit{others}. Each news sample is represented as an image-text pair, equipped with detailed manual annotation. The average length of manual annotations is 69.1 tokens. The overall manipulation rate of HFFN is 68.4\%, including 7.8\% of multi-modal manipulation samples.

\section{M-DRUM: Multi-modal news Detection and Reasoning langUage Model}

To address fake news detection and manipulation reasoning, as illustrated in Fig.\ref{fig:architecture}, we present M-DRUM, a novel large vision-language model based architecture. In M-DRUM, we use a multi-modal encoder to extract visual and textual features from news images and headlines. We leverage a cross-attention mechanism to obtain multi-modal fusion features. A prompt learner bridges the gap between manipulation expertise and the general knowledge of LVLM and based on that, a LVLM generates the analytical reasoning. The model is trained under a two-stage framework to strengthen the capacity of identification and reasoning. 

\subsection{Multi-modal Feature Extraction}

Driven by the idea of multi-modal alignment, we use ImageBind~\cite{girdhar2023imagebind}, a powerful cross-modal alignment model as the feature encoder. Given the news image $I\in\mathbb{R}^{H\times W\times C}$ and the corresponding headline $T$, we firstly extract visual and textual features with ImageBind. Inspired by AnomalyGPT~\cite{gu2023anomalygpt}, we obtained 4 intermediate visual features $F_{image}^i\in\mathbb{R}^{H_i\times W_i\times C_{image}}$ from encoding at different depths, where $i$ indicates the $i$-th depth. Accordingly, the textual feature $F_{text}\in\mathbb{R}^{C_{text}}$ is extracted from the headline.

To thoroughly comprehend multi-modal inputs, a cross-modal fusion is adopted to integrate uni-modal features. Focusing on the cross-modal relationship, the cross-modal features $F_{cross}^i\in\mathbb{R}^{C_{fusion}}$ are obtained by calculating the cross-attention score between the rectified visual feature and the textual feature in the softmax-free linear attention ~\cite{koohpayegani2024sima} expressions. The cross-modal fusion process can be represented as:
\begin{equation}
    \footnotesize
    \begin{aligned}
    \tilde{F}_{image}^i=&{\rm LinearLayer}(F_{image}^i), \\
    F_{cross}^i=&\tilde{F}_{image}^i\cdot F_{text}^T,
    \end{aligned}
\end{equation}
where $i$ indicates the $i$-th stage.

\noindent\textbf{Face encoder} In our work, human-centric news is highlighted which suffers from face swap and malicious editing. Therefore, identification aids provided by facial authenticity features are necessary. In M-DRUM, we leverage a face encoder to extract the manipulation-specific representation of human faces. The face encoder is modeled with a ResNet-50~\cite{he2016deep} and pretrained on large-scale Deepfake dataset~\cite{rossler2019faceforensics++} to provide feature-level guidance for identifying facial authenticity. The extracted facial feature $F_{face}$ is then concatenated with the cross-modal features to obtain the ultimate fusion feature. The fusion process can be represented as:
\begin{equation}
    F_{fusion} = {\rm Concate}(F_{cross}^i, F_{face}).
\end{equation}

\subsection{Manipulation Reasoning}

In M-DRUM, a LVLM serves as a mighty knowledge base for reasoning generation. To prompt LVLM with the authenticity of the news and inspire the general knowledge, we design a hybrid prompt learner to bridge the gap between the manipulation expertise and the general knowledge of LVLM. The prompt learner aims to assist the LVLM in understanding the manipulation information of multi-modal news comprehensively. As shown in Fig.\ref{fig:architecture}, the prompt learner integrates three parts of information. The fusion feature $F_{fusion}$ is transformed into prompt embeddings with a converter. A prediction head is introduced to provide specific guidance to the conversion process, supervised by the authenticity label of the news, together with bounding box of the edited regions. We expect the LVLM to accept semantic information from news images as much as possible so that it can be combined with facts for reasoning. To better transform the semantics information, we leverage a semantic learner to derive visual semantics from the visual features $F_{image}$ in the form of prompt embeddings. Additionally, to learn specific prompts for manipulation reasoning in a self-adaptive way, learnable prompt embeddings are adopted. The multi-level prompts are fed to the LVLM to raise analytical reasoning about potential manipulations. The global function of the prompt learner is:
\begin{equation}
    E_{prompt} = {\rm Concate}(C(F_{fusion}|H(F_{fusion})), L(F_{image}), E_{ada}),
\end{equation}
where $C$, $H$, $L$ stand for the converter, the prediction head and the semantic learner respectively. $E_{ada}$ stands for the self-adaptive prompt embeddings. 

\subsection{Loss Functions for Performance Augmentation}

To augment fake news detection and reasoning, we constrain our model with three types of loss functions: cross-entropy loss, bounding box loss and GIoU loss.

\noindent\textbf{Cross-entropy Loss} Cross-entropy loss is widely used in classification and natural language generation tasks. For authenticity classification, cross-entropy loss is introduced to supervise the prediction head in the prompt learner, which is defined as:
\begin{equation}
    \mathcal{L}_{CE}=-\frac{1}{\mathcal{B}}\sum_{i=1}^\mathcal{B}[y_i\log(p_i)+(1-y_i)\log(1-p_i)],
\end{equation}
where $\mathcal{B}$ is the batch size, $y_i$ is the authenticity label of the $i$-th sample and $p_i$ is the probability for positive prediction.

For manipulation reasoning, cross-entropy loss quantifies the disparity between the generated reasoning and the target text sequence, which is defined as:
\begin{equation}
    \mathcal{L}_{LLM}=-\frac{1}{n}\sum_{i=1}^ny_i\log(p_i),
\end{equation}
where $n$ is the number of tokens, $t_i$ is the ground truth label for token $i$ and $q_i$ is the predicted probability for token $i$.

\noindent\textbf{Bounding box Loss} We utilize L1 loss to supervise manipulated regions predicted by the prediction head, which is defined as:
\begin{equation}
    \mathcal{L}_{bbox}=\frac{1}{\mathcal{B}}\sum_{i=1}^\mathcal{B}|b_{pred}-b_{gt}|,
\end{equation}
where $\mathcal{B}$ is the batch size, $b_{pred}$ and $b_{gt}$ is the predicted and true bounding boxes respectively.

\noindent\textbf{GIoU Loss} Intersection over Union(IoU) loss is commonly used in object detection tasks with scale invariance. GIoU~\cite{rezatofighi2019generalized} serves as an improvement of IoU by optimizing in the case of non-overlapping bounding boxes, which is defined as:
\begin{equation}
    \mathcal{L}_{GIoU}=1-{\rm GIoU}=1-({\rm IoU} - \frac{|A^c-\mathcal{U}|}{A^c}),
\end{equation}
where $A^c$ and $\mathcal{U}$ is the smallest enclosing box and the union area of the predicted and the true bounding boxes respectively. We introduce bounding box loss and GIoU loss to assist our model in locating manipulated regions and understanding visual semantics.

The global loss function can be calculated by the weighting of each loss functions:
\begin{equation}
    \mathcal{L}=\alpha\mathcal{L}_{CE}+\beta\mathcal{L}_{LLM}+\gamma\mathcal{L}_{bbox}+\delta\mathcal{L}_{GIoU},
\end{equation}
where $\alpha, \beta, \gamma, \delta$ are hyper-parameters.

\subsection{Two-Stage Training Process}

To better combine the capabilities of multi-modal feature extraction and manipulation reasoning, we adopt a two-stage framework to train M-DRUM, refers to detection learning and reasoning learning.

\noindent\textbf{Detection Learning} In the detection learning stage, we set the face encoder and the prompt learner to be trainable. We train our model on the large-scale multi-modal media manipulation dataset ${\rm DGM}^4$~\cite{shao2023detecting}. During the training process, we expect the model to improve the performance of authenticity classification in large-scale detection task, which serves as the basis for subsequent manipulation reasoning.

\noindent\textbf{Reasoning Learning} In the reasoning learning stage, only the prompt learner is trainable. The training is launched on a delicately annotated human-centric and fact-related fake news detection benchmark HFFN. At this stage, we expect our model to improve analysis and reasoning abilities with multi-level prompts and generate analytical reasoning with confidence and vividness.

\begin{table*}
\begin{center}
\begin{tabular}{p{1.55cm}|p{0.8cm}<{\centering}|p{0.8cm}<{\centering}p{0.8cm}<{\centering}p{0.8cm}<{\centering}|p{0.8cm}<{\centering}p{0.7cm}<{\centering}p{0.8cm}<{\centering}|p{0.8cm}<{\centering}p{0.7cm}<{\centering}p{0.8cm}<{\centering}|p{0.8cm}<{\centering}p{0.7cm}<{\centering}p{0.8cm}<{\centering}}
    \hline
    \multirow{2}*{Method} & \multirow{2}*{Acc.} & \multicolumn{3}{c|}{Entertainment} & \multicolumn{3}{c|}{Sport} & \multicolumn{3}{c|}{Politics} & \multicolumn{3}{c}{Others} \\
    & & Pre. & Rec. & F1 & Pre. & Rec. & F1 & Pre. & Rec. & F1 & Pre. & Rec. & F1 \\
    \hline
    MCAN      & 0.710 & 0.896 & 0.729 & 0.804 & 0.892 & 0.776 & 0.830 & 0.811 & 0.652 & 0.723 & 0.850 & 0.667 & 0.747 \\
    HAMMER    & 0.722 & 0.951 & 0.639 & 0.765 & 0.913 & \textbf{0.913} & \textbf{0.913} & 0.818 & 0.643 & 0.720 & \textbf{0.914} & 0.653 & 0.762 \\
    \textbf{M-DRUM} & \textbf{0.804} & \textbf{0.980} & \textbf{0.820} & \textbf{0.893} & \textbf{0.951} & 0.853 & 0.899 & \textbf{0.872} & \textbf{0.810} & \textbf{0.840} & 0.913 & \textbf{0.840} & \textbf{0.875} \\
    \hline
\end{tabular}
\end{center}
\caption{Comparison among multi-modal fake news detection models on HFFN. The best-performing model is in \textbf{bold}.}
\label{tab:quantitative}
\end{table*}

\begin{table}
\begin{center}
\begin{tabular}{p{1.6cm}|p{1.2cm}<{\centering}p{1.2cm}<{\centering}p{1.2cm}<{\centering}|p{1.2cm}<{\centering}}
    \hline
    \multirow{2}*{Method} & \multicolumn{3}{c|}{Human Evaluation} & \multirow{2}*{Total} \\
    & Exactness & Certainty & Detail & \\
    \hline
    GT       & \textbf{9.30} & \textbf{8.97} & \textbf{8.48} & \textbf{8.92} \\
    \hline
    PandaGPT          & 1.42 & 4.65 & 1.82 & 2.63 \\
    GPT-4             & 2.33 & 2.80 & 4.60 & 3.24 \\
    LLaVA             & 2.03 & 7.17 & \textbf{7.40} & 5.53 \\
    \textbf{M-DRUM}   & \textbf{9.10} & \textbf{8.45} & 7.25 & \textbf{8.27} \\
    \hline
\end{tabular}
\end{center}
\caption{Human evaluation on the manipulation reasoning proposed by M-DRUM and powerful LVLMs(scores range from 1$\sim$10). The best-performing model is in \textbf{bold}. (GT: human annotation)}
\label{tab:human}
\end{table}

\begin{table}
\begin{center}
\begin{tabular}{p{0.9cm}|p{1.3cm}<{\centering}p{1.3cm}<{\centering}p{1.3cm}<{\centering}p{1.3cm}<{\centering}}
    \hline
    Setup & Accuracy & Precision & Recall & F1-score \\
    \hline
    0-shot & 0.804          & \textbf{0.934} & 0.833 & 0.880 \\
    1-shot & 0.396          & 0.647          & 0.253 & 0.364 \\
    2-shot & 0.596          & 0.810          & 0.615 & 0.699 \\
    4-shot & \textbf{0.863} & 0.885          & \textbf{0.968} & \textbf{0.925} \\
    \hline
\end{tabular}
\end{center}
\caption{Few-shot fake news detection results on HFFN. The best performance is highlighted in \textbf{bold}.}
\label{tab:fewshot}
\end{table}

\section{Experiments}

In this section, we evaluate the performance of M-DRUM on HFFN benchmark. Quantitative results on authenticity classification demonstrate M-DRUM outperforms SOTA multi-modal fake news detection models. The effectiveness of few-shot learning and chain-of-thought reasoning is also verified. Qualitative analysis on manipulation reasoning exhibits that M-DRUM proposes reasoning with more confidence and vividness compared with mainstream LVLMs.

\subsection{Experimental Settings}

\noindent\textbf{Baselines} Quantitative and qualitative experiments are performed on fake news detection and manipulation reasoning, respectively. For fake news detection, baselines are MCAN~\cite{wu2021multimodal} and HAMMER~\cite{shao2023detecting}, which are the state-of-the-art multi-modal detection models. For manipulation reasoning, we compare our method with powerful LVLMs including PandaGPT, GPT-4 and LLaVa by prompting them to propose reasoning on potential media manipulations.

\noindent\textbf{Metrics} In quantitative experiments, we evaluate models on accuracy, precision, recall and F1-score, which are commonly used in fake news detection tasks to measure the performance of authenticity classification. In qualitative experiments, we evaluate the reasoning results manually. 12 independent human raters are employed to assess the quality of randomly chosen reasoning results. Human evaluation is conducted on three orthogonal aspects: \textit{Exactness}, \textit{Certainty} and \textit{Detail}. \textit{Exactness} refers to whether the reasoning results are correct and consistent with the news content. \textit{Certainty} refers to whether the reasoning results are clear and an ambiguous answer is regarded as a low score. \textit{Detail} refers to whether the reasoning results are analyzed in detail rather than talk in generalities. Human raters are asked to score the reasoning on a scale of 1 to 10, with higher scores indicating better performance. We show the average evaluation scores on three aspects of M-DRUM and mainstream LVLMs.

\noindent\textbf{Implementation Details} We use ImageBind-Huge~\cite{girdhar2023imagebind} as the multi-modal feature encoder and Vicuna-7B\cite{vicuna2023} as the LVLM. We concatenate the outputs from the 8th, 16th, 24th, 32nd layers of the image encoder into the visual feature. The parameters of the model are initialized using the pre-trained parameters in PandaGPT~\cite{su2023pandagpt}. In both the detection and reasoning learning stage, training is performed with a learning rate of 1e-3 and a batch size of 16. We set the image resolution to be 224$\times$224. Loss weights $\alpha, \beta, \gamma, \delta$ are set to 1 by default. All of our experiments are conducted on 2 NVIDIA 3090 GPUS with PyTorch framework.

\begin{figure}[t]
    \centering
     \begin{minipage}[t]{0.47\linewidth}
        \centering
        \includegraphics[height=100pt]{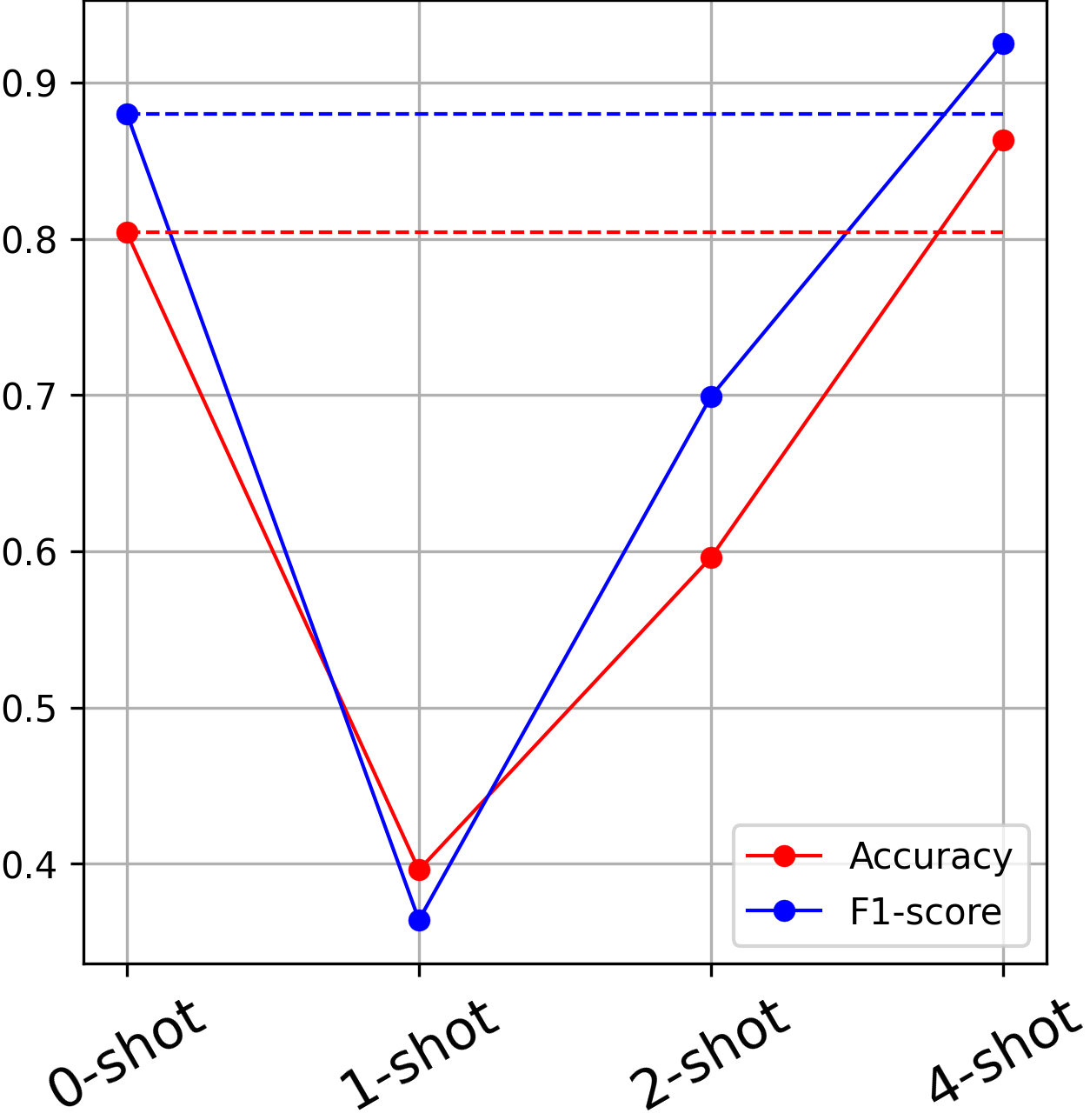}
        \caption{Performance floating of few-shot learning.}
        \label{fig:fewshot}
    \end{minipage}
    \begin{minipage}[t]{0.47\linewidth}
        \centering
        \includegraphics[height=100pt]{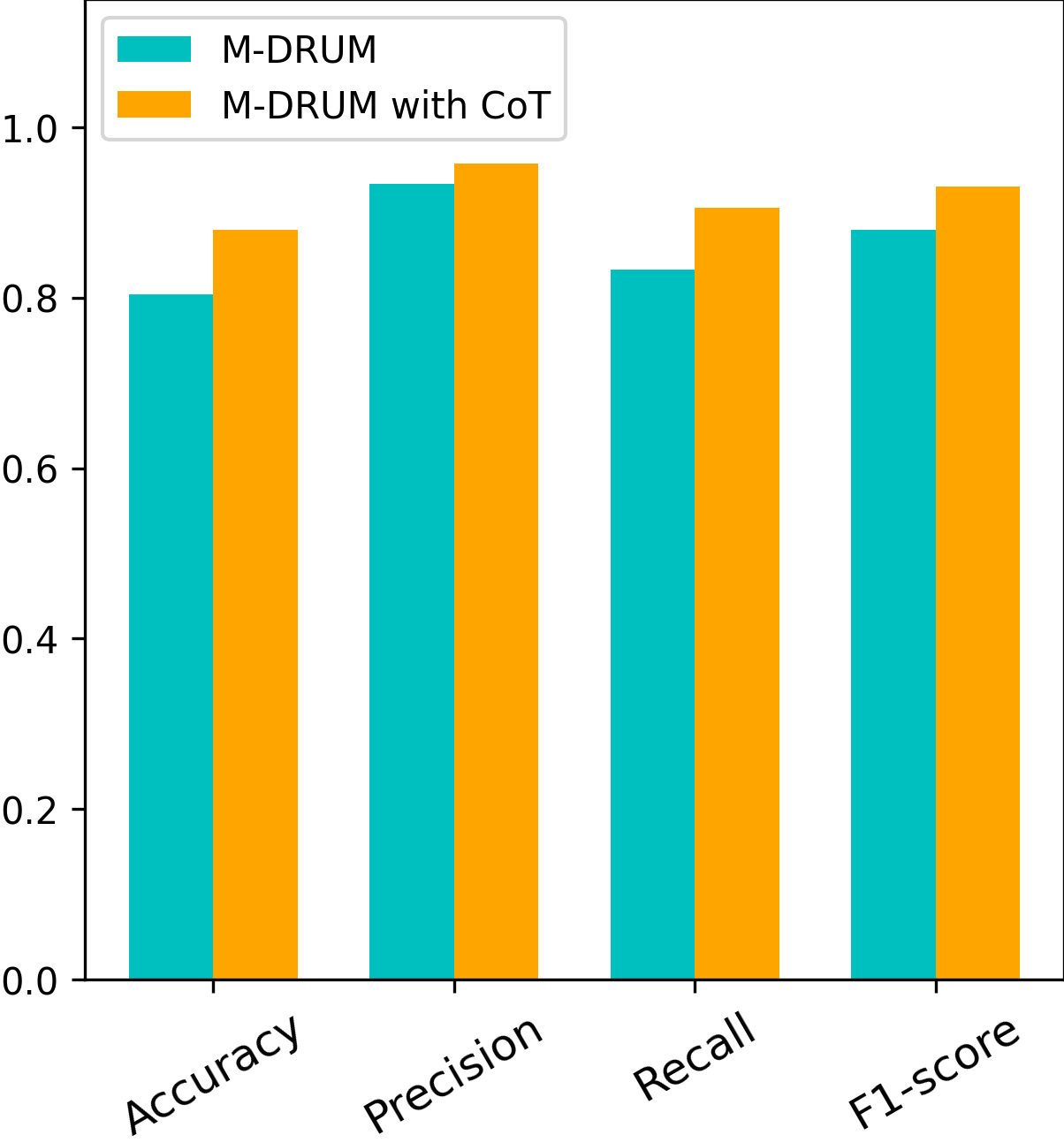}
    \caption{Efficacy of chain-of-thought (CoT) reasoning.}
    \label{fig:cot}
    \end{minipage}
\end{figure}

\begin{figure*}[ht]
    \includegraphics[width=1\textwidth]{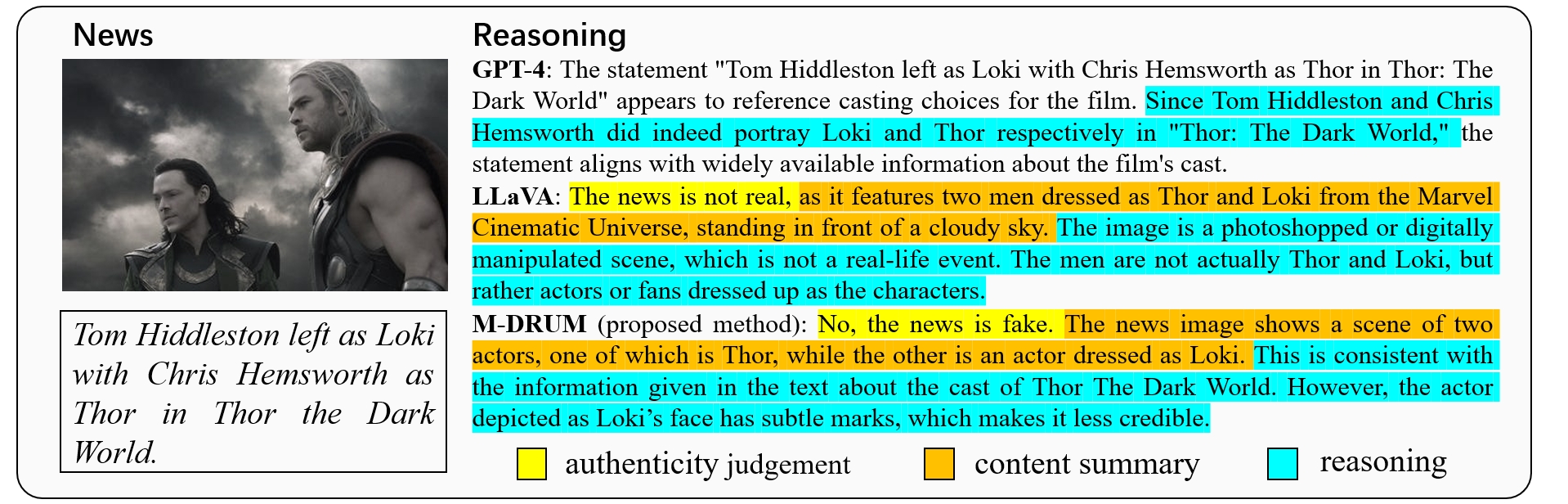}
    \caption{An example of the manipulation reasoning proposed by M-DRUM, compared with powerful LVLMs. The content analysis of the reasoning is marked with different colors.}
    \label{fig:qualitative}
\end{figure*} 

\begin{table*}
\centering
\begin{tabular}{p{2.5cm}|p{1.3cm}<{\centering}p{1.3cm}<{\centering}p{1.3cm}<{\centering}|p{1.3cm}<{\centering}p{1.3cm}<{\centering}p{1.3cm}<{\centering}p{1.3cm}<{\centering}}
    \hline
    Method & Image & Text & Face & Accuracy & Precision & Recall & F1-score \\
    \hline
    M-DRUM       & \checkmark & \checkmark & \checkmark & \textbf{0.804} & \textbf{0.934} & \textbf{0.833} & \textbf{0.880} \\
    M-DRUM w.o I & & \checkmark & \checkmark            & 0.569 & 0.839 & 0.580 & 0.686 \\
    M-DRUM w.o T & \checkmark & & \checkmark            & 0.388 & 0.571 & 0.475 & 0.519 \\
    M-DRUM w.o F & \checkmark & \checkmark &            & 0.745 & 0.890 & 0.794 & 0.840 \\
    \hline
\end{tabular}
\caption{Results of ablation studies on modalities of M-DRUM. The \checkmark indicates module inclusion.}
\label{tab:ablation}
\end{table*}

\subsection{Authenticity Classification}

We compare M-DRUM with SOTA multi-modal fake news detection models MCAN and HAMMER on HFFN benchmark. The results are presented in Tab.\ref{tab:quantitative}. By comparison, M-DRUM exceeds the performance of the baselines on HFFN benchmark. Specifically, M-DRUM achieves the highest accuracy of 80.4\%, exceeding the SOTA fake news detection models by 8.2\%. In four domains of HFFN, M-DRUM ranks either first or second in terms of precision, recall and F1-score. Quantitative results exhibits the superiority of M-DRUM to identify the human-centric and fact-related news. The performance advantage of M-DRUM on HFFN highly relies on the emphasis of facial feature and the general knowledge owned by the LVLM to which feature-based detection models are not comparable.

\noindent\textbf{Few-shot Learning} Considering the implementation of LVLM, we expect the M-DRUM to perform better under few-shot learning~\cite{brown2020language} settings. We test the performance of M-DRUM in 0, 1, 2 and 4-shot learning settings. The result is shown in Tab.\ref{tab:fewshot} and Fig.\ref{fig:fewshot}. As the prompt examples adding, the classification performance of M-DRUM declines and then climb up. This can be explained with: a small number of prompt examples tend to confuse the model, and a larger number of samples promote the model to synthesize implicit rules and make correct judgements. Furthermore, 4-shot learning outperforms 0-zero by 5.9\% in accuracy, indicating an appropriate number of examples leads to better performance, which verifies the promotion of few-shot learning.

\noindent\textbf{Chain-of-Thought Reasoning} A chain-of-thought (CoT) refers to a series of intermediate reasoning steps that mimic the reasoning process of human and significantly promote LLMs to tackle complex tasks~\cite{wei2022chain}. In HFFN, manual annotations of \textit{content summary} and \textit{clue revelation} serves as the intermediate steps of the CoT. We explore whether the reasoning ability of M-DRUM can be improved through a step-by-step process in few-shot learning. In the CoT strategy, the \textit{content summary} and the \textit{clue revelation} of news are added to training examples and M-DRUM is guided to reason based on both original news content and manually annotated intermediate steps. Fig.\ref{fig:cot} demonstrates the performance of CoT reasoning, where M-DRUM receives a 7.6\% accuracy boost and a 5.0\% F1-score boost with CoT instruction. The result shows that generating the CoT along with the answer benefits the detection of M-DRUM.

\subsection{Manipulation Reasoning}

We encourage the detection model to propose analytical reasoning about manipulation in assist of unveiling forgery mechanisms. To evaluate the reasoning results, we scored M-DRUM and mainstream LVLMs on a manual basis. Tab.\ref{tab:human} shows the result of human evaluation. In terms of \textit{exactness} and \textit{certainty}, M-DRUM far exceeds other LVLMs. Slightly inferior to LLaVA, reasoning results proposed by M-DRUM are still rich in \textit{detail}. In general, the analytical reasoning proposed by M-DRUM can reach the performance of the ground truth (manual annotations). An example of the manipulation reasoning is shown in Fig.\ref{fig:qualitative}. Compared with the detailed manipulation reasoning generated by M-DRUM, the reasoning of GPT-4 is ambiguous and the reasoning of LLaVA is multi-leveled but flawed. Conclusively, M-DRUM raises analytical reasoning with more confidence and vividness, which greatly expands the implementation of fake news detection models.

\subsection{Ablation Studies}

To evaluate the role of each modality in fake news detection and verify the effectiveness of architecture design, ablation experiments are conducted. We explore the impact of ignoring visual, textual and facial features in detection. In each set of experiment, a certain modality of M-DRUM is eliminated by removing the corresponding feature encoder and the two-stage training process is re-conducted. We collated the classification results of the ablation model which are presented in Tab.\ref{tab:ablation}. It can be observed that all ablation models with certain modality eliminated suffers from severe performance degradation, which proves the importance of each modality in M-DRUM and the necessity of facial authenticity features toward human-centric news detection.

\section{Conclusion}

In this paper, we introduce the fake news detection and manipulation reasoning benchmark HFFN. The benchmark is constructed following the principles of "\textit{human-centric}" and "\textit{fact-related}". To address classification and reasoning, we present M-DRUM, a novel detection model leveraging LVLM as the backbone. Combining multi-modal manipulation expertise and the general knowledge of LVLM, M-DRUM can not only perform authenticity judgement on the multi-modal news, but also enable analytical reasoning about potential manipulations. Comprehensive experiments demonstrate that M-DRUM outperforms SOTA fake news detection models and mainstream LVLMs. 
Further experiments verify the improvement of few-shot learning and chain-of-thought reasoning. Ablation studies exhibit the indispensability of different modals in detection.


\bibliographystyle{ACM-Reference-Format}
\bibliography{sample-base}










\end{document}


\title{Supplementary Materials: Fake News Detection and Manipulation Reasoning via Large-Vision Language Models}


\author{Anonymous Authors}

\maketitle

\section{The Data Format of HFFN Benchmark}

In our work, We propose the benchmark of \textbf{H}uman-centric and \textbf{F}act-related \textbf{F}ake \textbf{N}ews (\textbf{HFFN}). The data format of HFFN are illustrated in Tab.\ref{tab:format} with descriptions. Encompassing four domains and five manipulation types, news with both image and text modalities was presented with detailed manual annotations. By facilitating the benchmark with detailed human annotations, we expect to leverage HFFN for evaluating the performance on authenticity classification and analytical reasoning of fake news detection models and general-purposed LVLMs.

\section{Implementation of the Baseline Methods}

As the supplements to \textit{section 5.1}, all the baseline methods were implemented with their publicly available source codes on 2 NVIDIA 3090 GPUS with PyTorch framework. We compare our model with SOTA detection models including MCAN and HAMMER in fake news detection task. For MCAN, we pre-trained it on large-scale multi-modal media manipulation dataset ${\rm DGM}^4$ and carefully tuned it on HFFN to achieve optimal performance. For HAMMER, we utilized the default settings provided in the original paper. In manipulation reasoning task, our model is compared with mainstream LVLMs including PandaGPT, GPT-4 and LLaVA. Specifically, the open-source parameters of \verb|pandagpt_7b_max_len_512| and \verb|llava-v1.5-7b| were used to launch PandaGPT and LLaVA respectively. We designed over 10 prompt templates for manipulation reasoning respectively and pre-test is conducted to evaluate which template leads to better performance on manipulation reasoning. Considering LVLMs are large-scale black-box models, we evaluate the potential of the templates from the quality of the inference results. The template for the best results is chosen as the final prompt template. We present the prompts used in our experiments to evaluate the manipulation reasoning ability of M-DRUM as follows:

\noindent\textbf{Prompts for manipulation reasoning}
$<$Img$>E_{img}<$/Img$>$ Assume you are an expert in manipulation reasoning. This is a photo selected from a piece of news, which needs to be real and consistent with the headline of the news: \{headline\}. The news may be confronted with media manipulations. You are required to reason about manipulations of the news. The reasoning needs to be consistent to the news content, and to be clear and detailed. Reasoning result: \{answer\}

Prompts for PandaGPT, GPT-4 and LLaVA are similar to this with slightly differences. They are not listed here due to space constraints.

\section{Performance of Few-shot Learning}

\begin{figure*}[t]
    \subfigure[accuracy]{
    \label{fig:accuracy}
    \includegraphics[width=0.47\columnwidth]{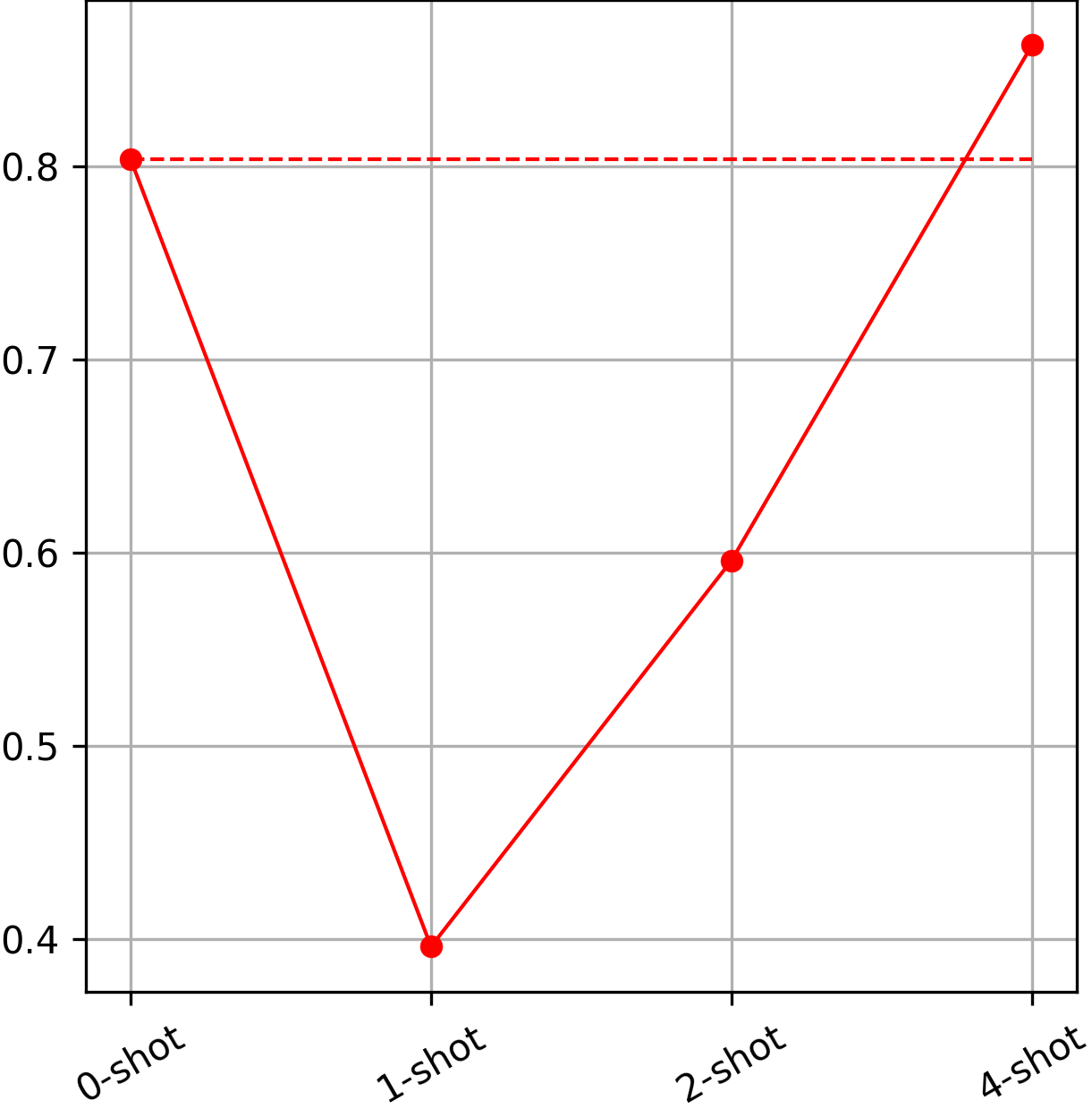}
    }
    \subfigure[precision]{
    \label{fig:precision}
    \includegraphics[width=0.48\columnwidth]{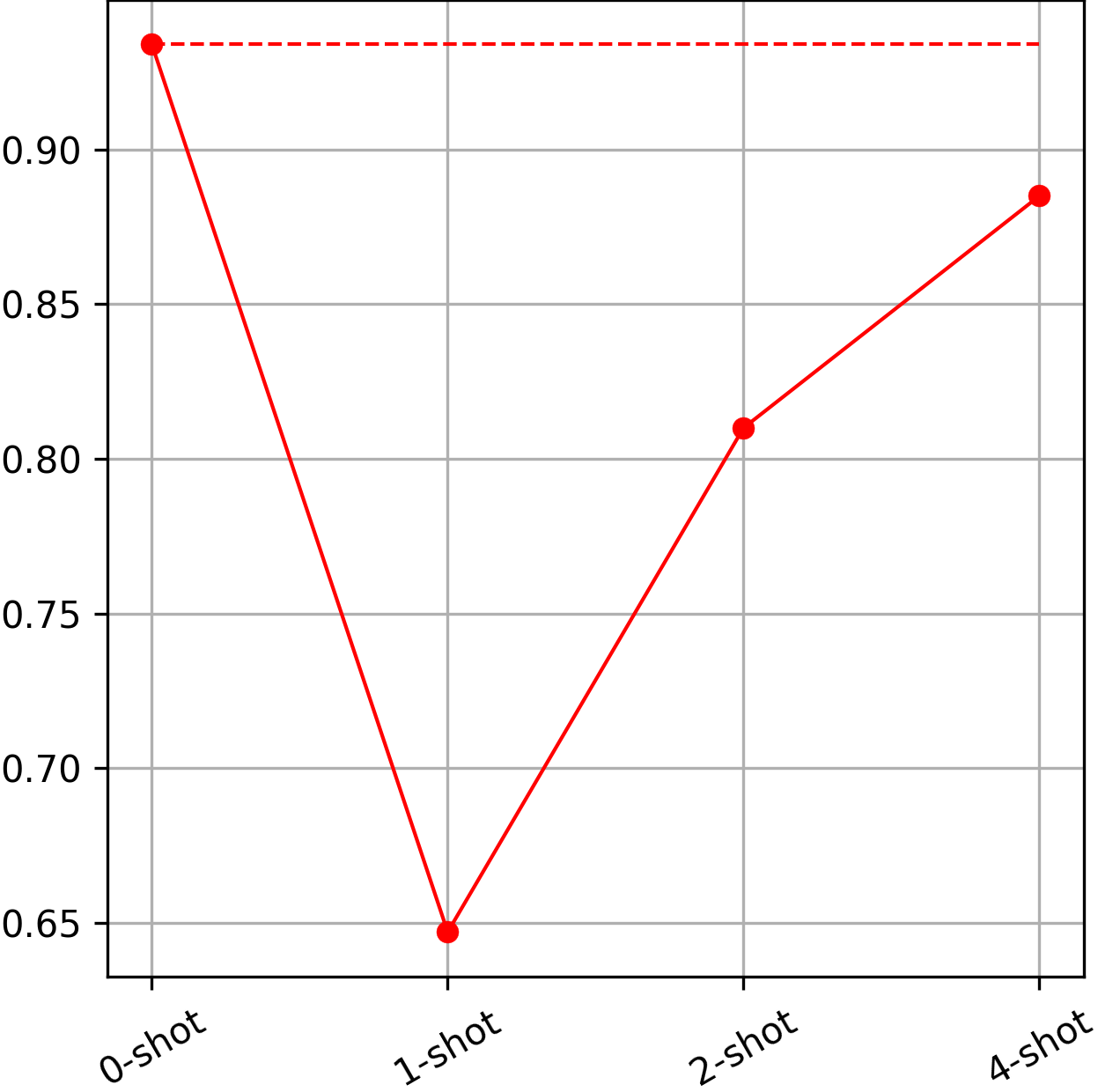}
    }	
    \subfigure[recall]{
    \label{fig:recall}
    \includegraphics[width=0.47\columnwidth]{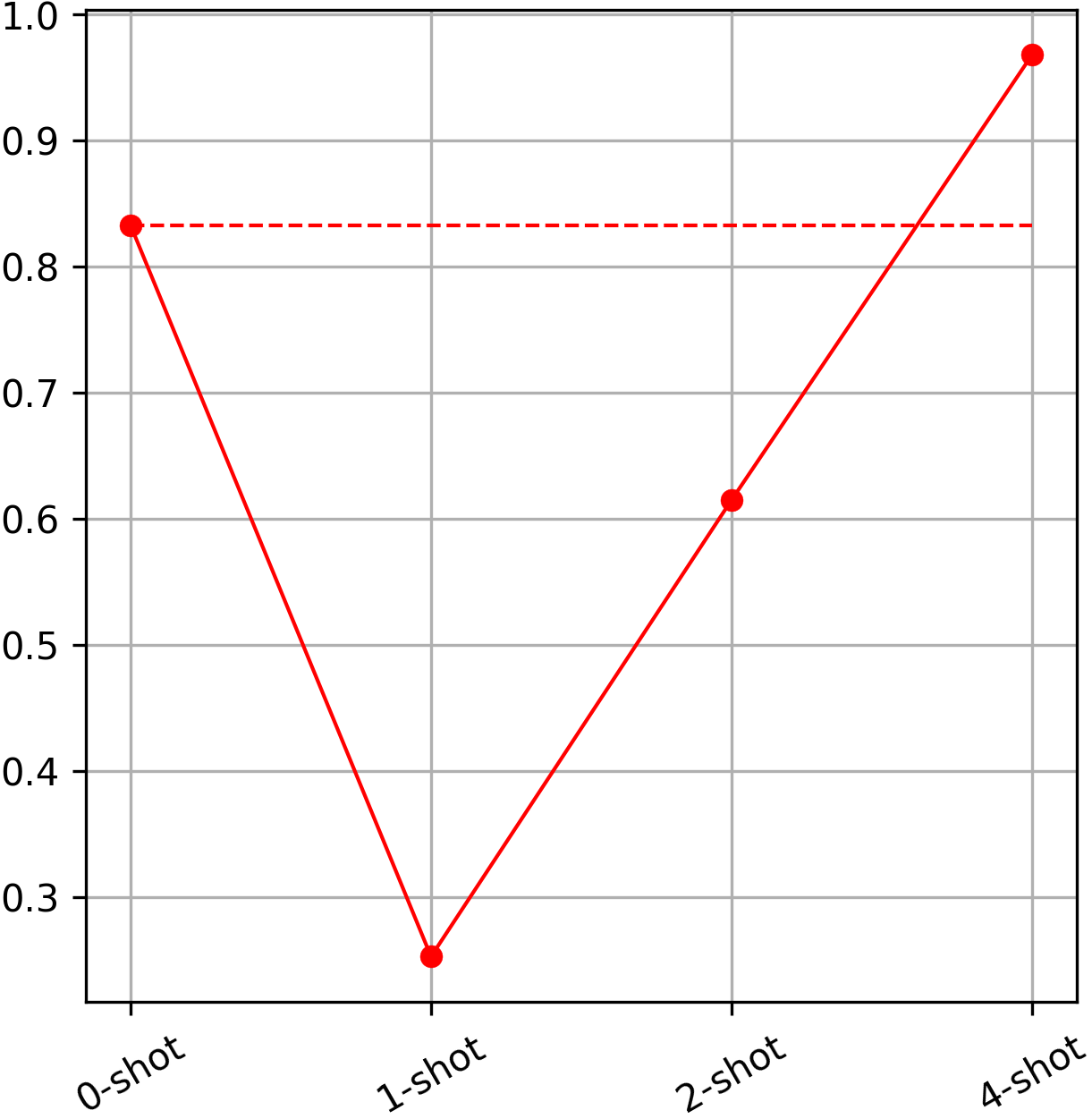}
    }
    \subfigure[f1-score]{
    \label{fig:f1-score}
    \includegraphics[width=0.47\columnwidth]{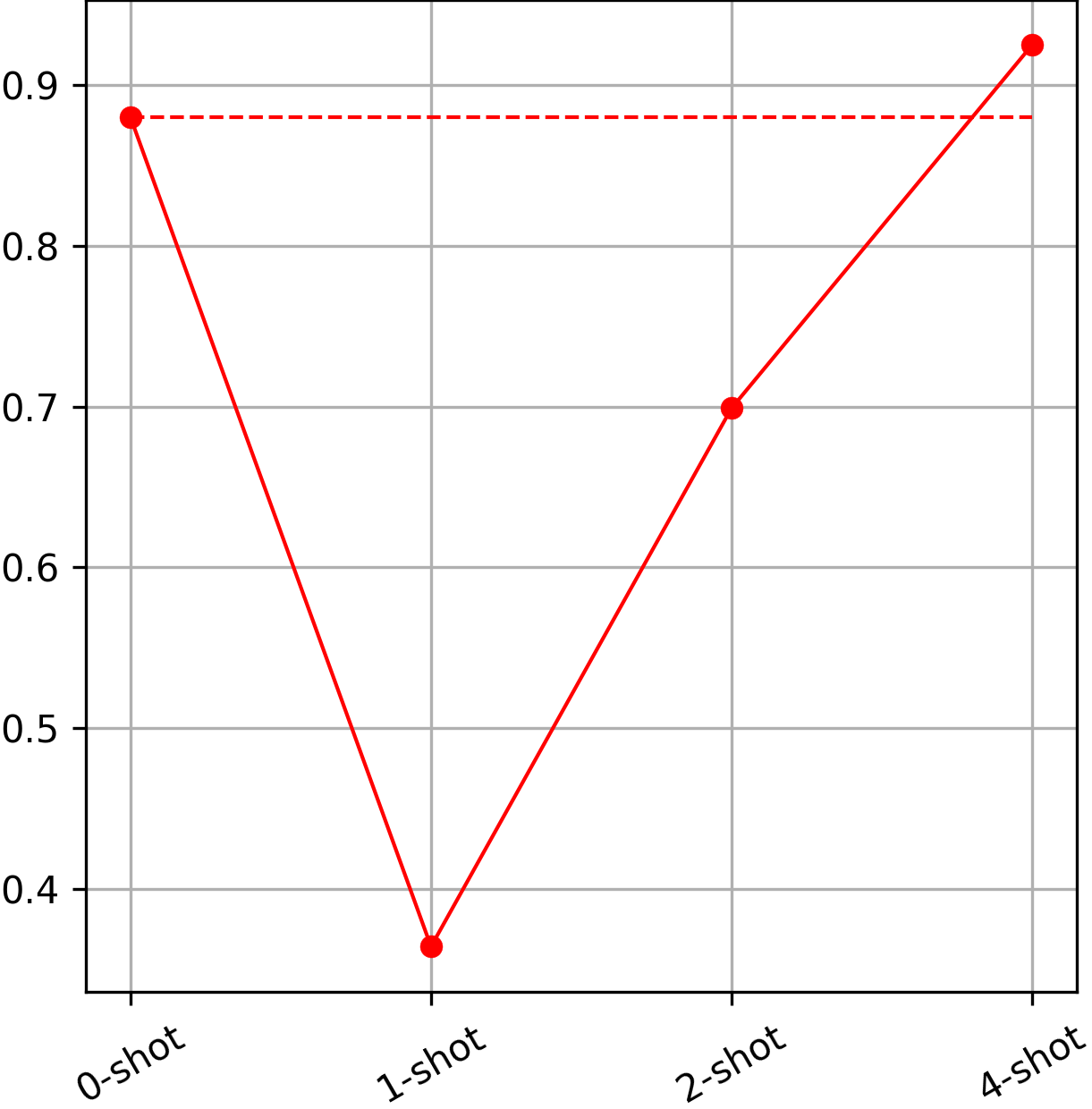}
    }
    \caption{Performance floating of few-shot learning.}
    \label{fig:fewshot}
\end{figure*}

The detailed inference results of M-DRUM under few-shot learning settings are illustrated in Fig.\ref{fig:fewshot}. Under the measurement of each indicators, the classification performance of M-DRUM declines and then climb up as the prompt examples adding.

\section{More Examples of Manipulation Reasoning}

\begin{figure*}[ht]
    \includegraphics[width=0.95\textwidth]{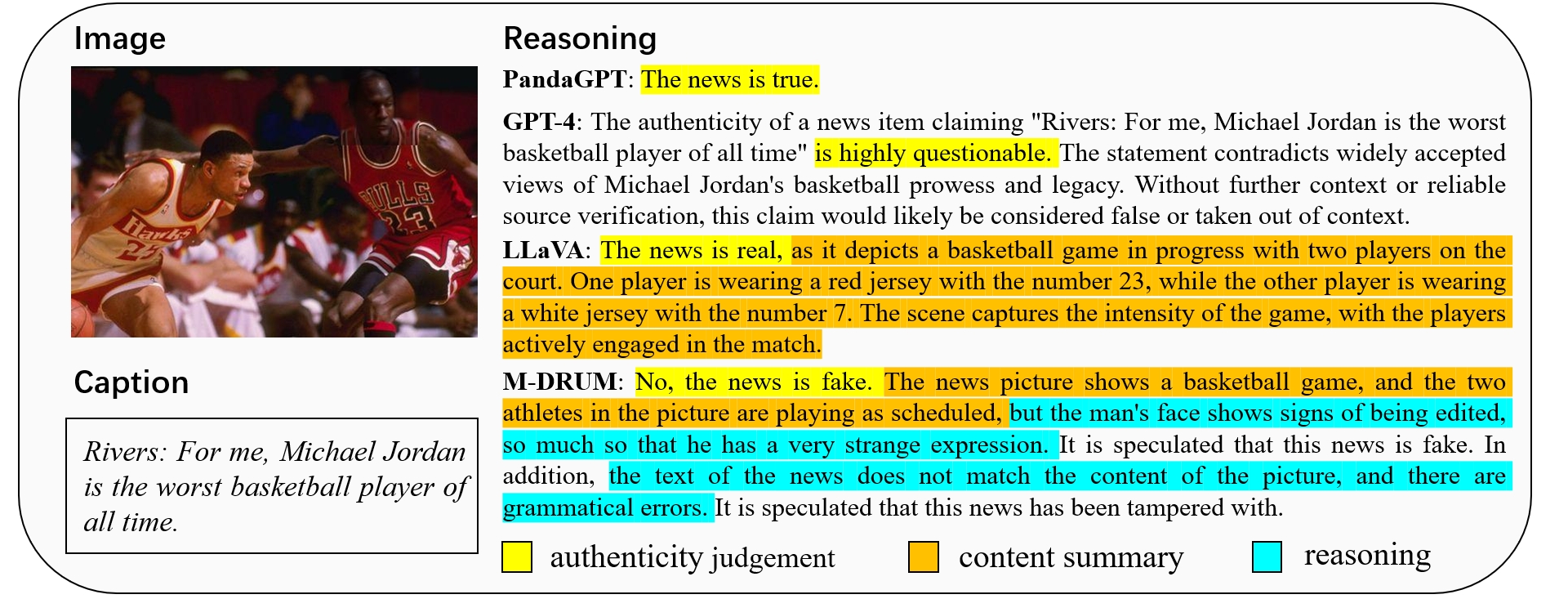}
    \caption{An example of manipulation reasoning in \textit{sport} domain, which is visually and textually manipulated.}
    \label{fig:example1}
\end{figure*}

\begin{figure*}[ht]
    \includegraphics[width=0.95\textwidth]{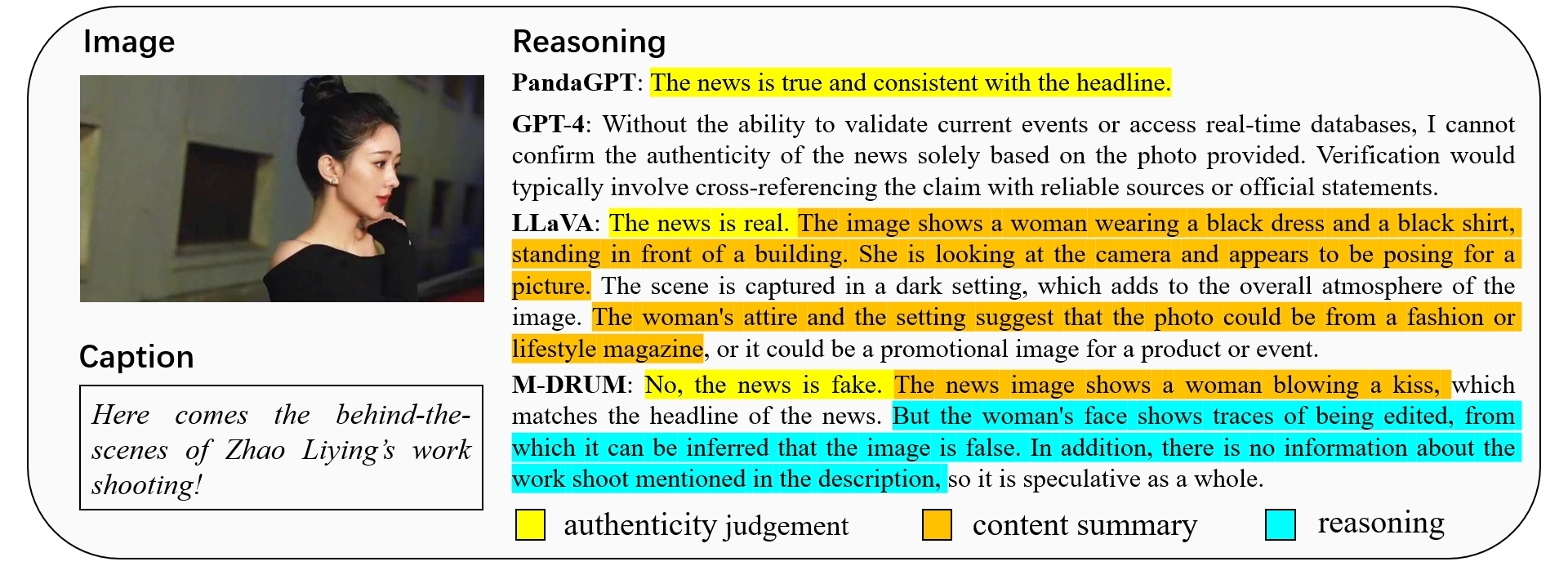}
    \caption{An example of manipulation reasoning in \textit{entertainment} domain, which is visually manipulated}
    \label{fig:example2}
\end{figure*}

\begin{figure*}[ht]
    \includegraphics[width=0.95\textwidth]{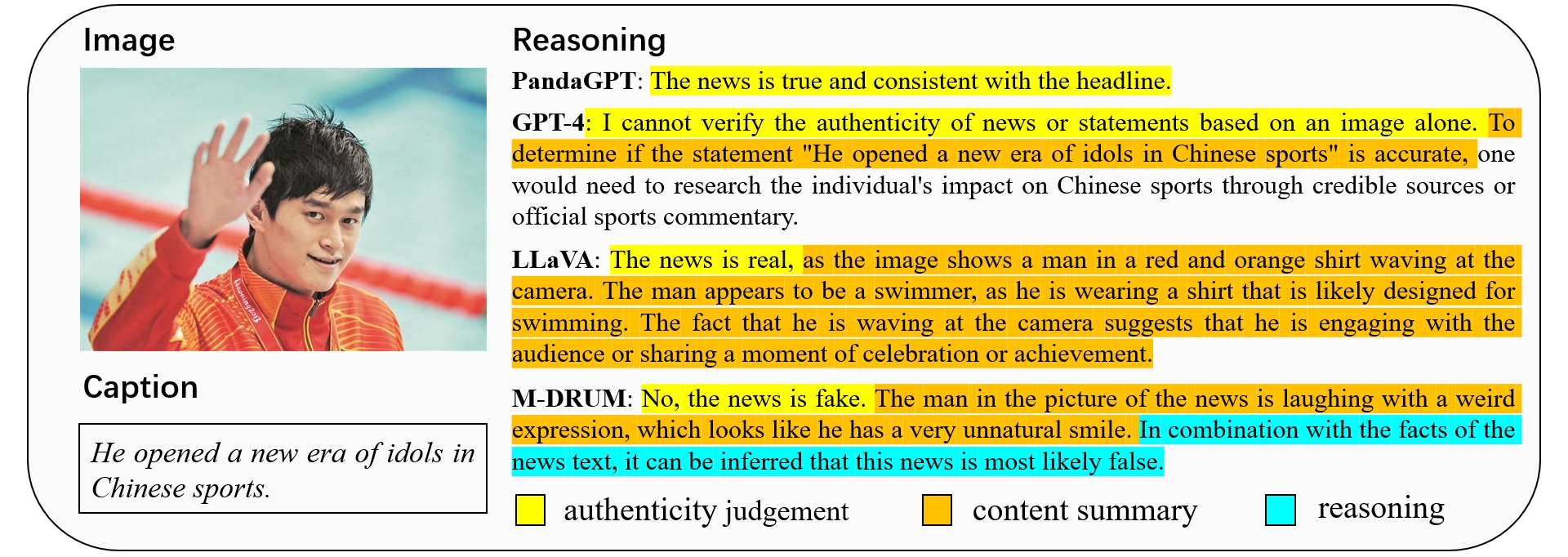}
    \caption{An example of manipulation reasoning in \textit{sport} domain, which is visually manipulated.}
    \label{fig:example3}
\end{figure*}

\begin{figure*}[ht]
    \includegraphics[width=0.95\textwidth]{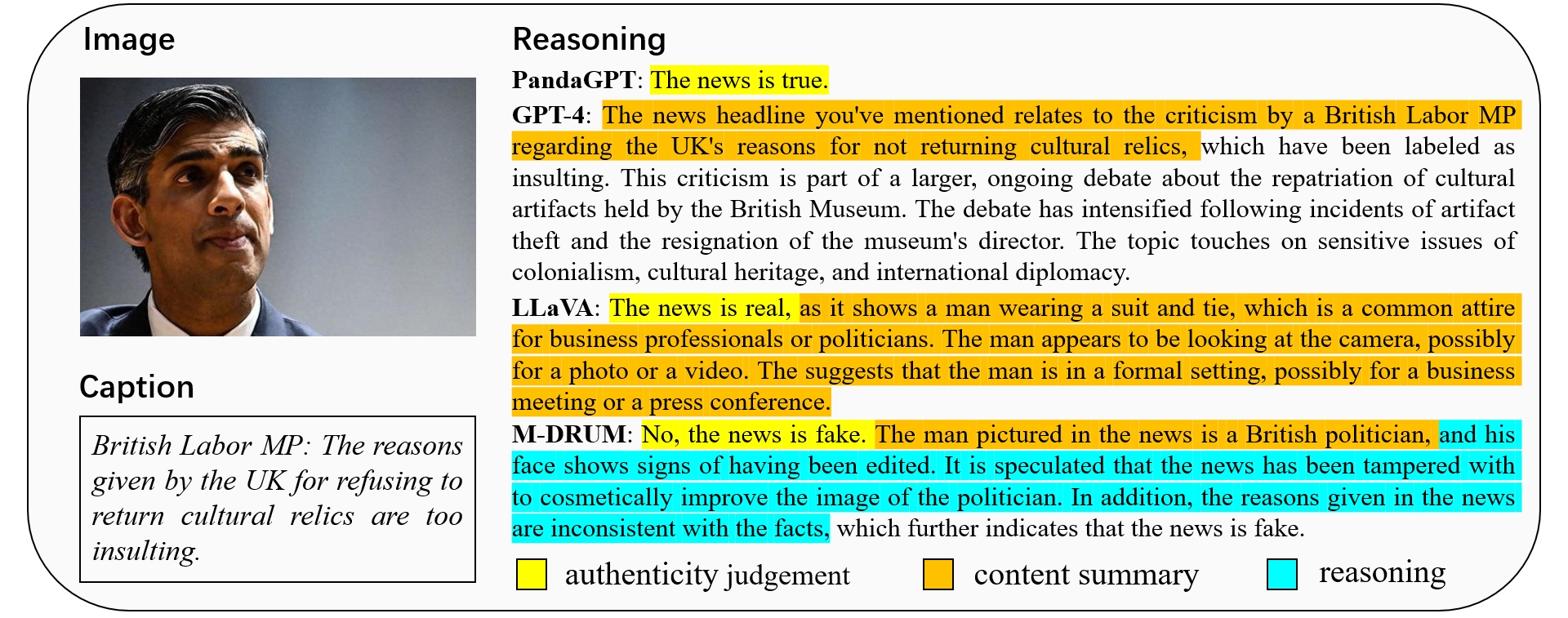}
    \caption{An example of manipulation reasoning in \textit{politics} domain, which is visually and factually manipulated.}
    \label{fig:example4}
\end{figure*}

\begin{figure*}[ht]
    \includegraphics[width=0.95\textwidth]{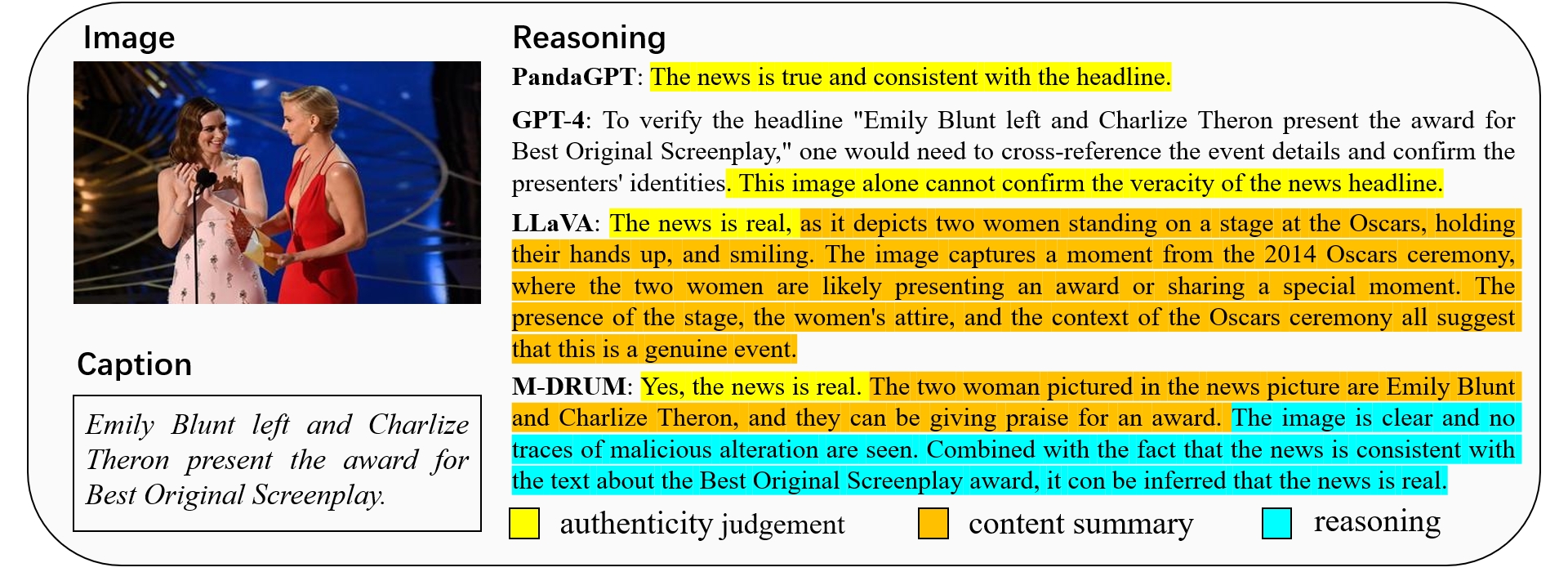}
    \caption{An Out-Of-Distribution example of manipulation reasoning in ${\rm DGM}^4$ dataset, which is real.}
    \label{fig:example5}
\end{figure*}

We compared our model with mainstream LVLMs in \textit{section 5.3}, specifically selecting PandaGPT, GPT-4 and LLaVA on manipulation reasoning. More examples of the comparative analysis on HFFN are presented in Fig.\ref{fig:example1}, Fig.\ref{fig:example2}, Fig.\ref{fig:example3} and Fig.\ref{fig:example4}. Outside the scope of HFFN benchmark, we conduct a small-scale test on the ${\rm DGM}^4$ dataset with M-DRUM and mainstream LVLMs. Fig.\ref{fig:example5} illustrates the out-of-distribution performance of M-DRUM on ${\rm DGM}^4$ dataset. It can be observed that PandaGPT frequently misjudges the authenticity or is unable to produce coherent words. In the absence of particular analysis, GPT-4 tends to propose ambiguous conclusions. Despite the in-depth explanation of the news content provided, LLaVA's analytical reasoning is not very accurate. While other models show unsatisfactory performance on manipulation reasoning, our model demonstrates proficiency in proposing reasoning combining the description of news content and the analysis on the potential manipulations. Furthermore, our model can generate analytical reasoning with more confidence and vividness, expanding the implementation of fake news detection models.

\begin{table}
\begin{center}
\begin{tabular}{|p{1.4cm}|p{6.4cm}|}
    \hline
    Keys &  \multicolumn{1}{c|}{Description} \\
    \hline
    \textbf{image}       & The image of news. \\
    \textbf{text}        & The headline of news. \\
    \hline
    \textbf{domain}      & The domain of news. (one of \textit{entertainment}, \textit{sport}, \textit{politics} and \textit{others})   \\
    \textbf{label}       & The authenticity label of news. (0 stands for real and 1 stands for fake.)   \\
    \textbf{fake\_cls}   & The manipulation class of news. (\textit{orig} for real news or one of the five manipulation types, including three uni-modal types and two cross-modal types.) \\
    \textbf{face\_bbox}  & The bounding box of the main human face, which is regarded as the fake region if the news image is manipulated. \\
    \hline
    \textbf{reasoning}  & Human annotated result of the analytical reasoning on manipulations. \\
    \hline
\end{tabular}
\end{center}
\caption{Data formats of HFFN benchmark.}
\label{tab:format}
\end{table}

\section{Details of Human Evaluation}

In human evaluation, each volunteer is paid to rate the results of different models' reasoning on five randomly selected items. All of our 12 raters are professional and diverse in background. The process of evaluation is independent with samples randomly chosen and shuffled for evaluation to reduce the impact of subjectivity. Specific human rating of different models on manipulation reasoning is exhibited in the supplementary Excel tables due to the space limitation.

